\documentclass[twoside,11pt]{article}

\usepackage{multirow}
\usepackage{booktabs}

\usepackage{bm}
\usepackage{algorithm}
\usepackage{algpseudocode}
\usepackage{amsmath}
\usepackage{amsfonts}
\algrenewcommand\algorithmicrequire{\textbf{Input:}}
\algrenewcommand\algorithmicensure{\textbf{Output:}}
\algrenewcommand\algorithmiccomment[1]{\hfill \textcolor{gray}{\textit{// #1}}}
\newcommand{\LineComment}[1]{\Statex \textcolor{gray}{\textit{// #1}}}
\usepackage{xcolor}

\usepackage{graphicx}
\usepackage{subcaption}

\usepackage{array}
\usepackage{booktabs}
\usepackage{enumitem}
\usepackage[abbrvbib, preprint]{ArXiv}
\usepackage[svgnames]{xcolor}
\hypersetup{
hidelinks,
colorlinks=true,
linkcolor=Indigo,
urlcolor=DeepSkyBlue4,
citecolor=Indigo
}
\usepackage{authblk}
\usepackage{amsmath} 

\usepackage{algpseudocode}
\renewcommand{\algorithmicrequire}{\textbf{Input:}}
\renewcommand{\algorithmicensure}{\textbf{Output:}}

\DeclareFixedFont{\myfont}{OT1}{ptm}{m}{n}{8pt}
\DeclareFixedFont{\myfontb}{OT1}{ptm}{bx}{n}{8pt}

\usepackage{marvosym}
\makeatletter
\def\@fnsymbol#1{\ifcase#1\or \text{\Letter}\or *\or \dagger\or \ddagger\else\@arabic{#1}\fi}
\makeatother

\usepackage{lastpage}
\jmlrheading{23}{\ Preprint 2026}{1-\pageref{LastPage}}{1/19}{2/5}{21-0000}{}

\firstpageno{1}

\begin{document}

\title{Enhance and Reuse: A Dual-Mechanism Approach to Boost Deep Forest for Label Distribution Learning}


\author[1,2]{Jia-Le Xu}
\author[1,2,3,4,\thanks{Corresponding author: lvsh@hhu.edu.cn}]{Shen-Huan Lyu}
\author[1,2]{Yu-Nian Wang}
\author[1,2]{Ning Chen}
\author[1,2]{Zhihao Qu}
\author[1,2]{Bin Tang}
\author[4]{Baoliu Ye}

\affil[1]{Key Laboratory of Water Big Data Technology of Ministry of Water Resources, Hohai University, Nanjing 211100, China}
\affil[2]{College of Computer Science and Software Engineering, Hohai University, Nanjing 211100, China}
\affil[3]{Department of Computer Science, City University of Hong Kong, Hong Kong 999077, China}
\affil[4]{National Key Laboratory for Novel Software Technology, Nanjing University, Nanjing 210023, China}

\editor{My editor}

\maketitle

\begin{abstract}
Label distribution learning (LDL) requires the learner to predict the degree of correlation between each sample and each label. To achieve this, a crucial task during learning is to leverage the correlation among labels. Deep Forest (DF) is a deep learning framework based on tree ensembles, whose training phase does not rely on backpropagation. DF performs in-model feature transform using the prediction of each layer and achieves competitive performance on many tasks. However, its exploration in the field of LDL is still in its infancy. The few existing methods that apply DF to the field of LDL do not have effective ways to utilize the correlation among labels. Therefore, we propose a method named \textbf{E}nhanced and \textbf{R}eused Feature \textbf{D}eep \textbf{F}orest (ERDF). It mainly contains two mechanisms: feature enhancement exploiting label correlation and measure-aware feature reuse. The first one is to utilize the correlation among labels to enhance the original features, enabling the samples to acquire more comprehensive information for the task of LDL. The second one performs a reuse operation on the features of samples that perform worse than the previous layer on the validation set, in order to ensure the stability of the training process. This kind of Enhance-Reuse pattern not only enables samples to enrich their features but also validates the effectiveness of their new features and conducts a reuse process to prevent the noise from spreading further. Experiments show that our method outperforms other comparison algorithms on six evaluation metrics. 
\end{abstract}


\section{Introduction}
In multi-label learning (MLL)~\citep{zhang2013review}, the label corresponding to an instance is a subset of labels, rather than a single label as in traditional single-label learning (SLL). For example, in movie classification, a movie can be labeled with ``comedy'', ``action'', and ``romance'' simultaneously; in news text annotation, an article can involve both ``technology'' and ``finance'' topics. MLL significantly enhances the model's ability to describe complex objects and has been widely applied in various fields such as text classification~\citep{ma2021label,wu2024contrastive}, multimedia analysis~\citep{lvsm,gnam} and image content annotation~\citep{chen2021learning,chencasual, ZHANG2026111959}.


Although MLL has significantly improved the expressive power compared to SLL, it shares the same fundamental assumption as SLL: labels describe instances in a binary manner, i.e., a label is either relevant or irrelevant for an instance. This binary assumption ignores a crucial fact that the contribution or importance of different labels in describing the same instance may vary from each other. For example, a movie might mainly be an action film but also contain a few comedy elements. Traditional learning paradigms cannot capture this semantic ambiguity and uncertainty. To more precisely quantify the relative importance of different labels, label distribution learning (LDL)~\citep{geng2016label} is proposed as a more general learning paradigm. The core idea of LDL is to associate a label distribution with an instance. This distribution is a real-valued vector, where each element represents the description degree of the corresponding label for the instance, and the sum of all dims is 1. In this way, LDL extends the qualitative ``yes or no'' judgment of label descriptions to a quantitative ``degree'' measurement, thereby enabling more precise and comprehensive capture of the semantic information of the object. Due to its ability to address the more general scenario of label ambiguity, it has seen wide application in diverse tasks such as facial expression recognition~\citep{chen2020facial, le2023facial, shin2024facial, khelifa2025facial}, object detection~\citep{xu2023object}, medical care~\citep{nishio2023cancer, shrivastava2023medical} and other related fields. Some recent articles innovatively use label distribution to assist in the process multi-label learning~\citep{chenning}.

\begin{figure}
\centering
\includegraphics[width=0.8\textwidth]{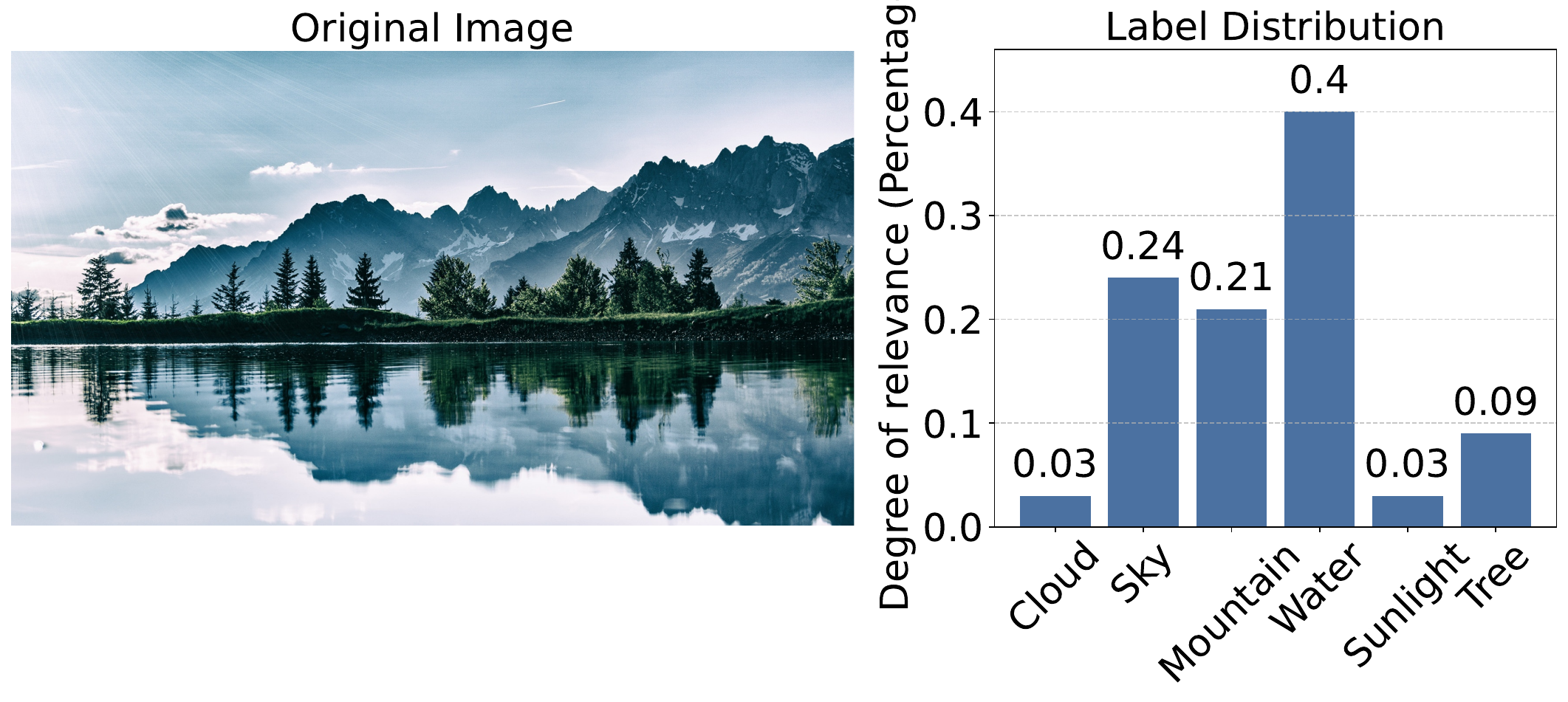}
\caption{This is an example sample of label distribution learning. On the left is a landscape picture as an input, which contains multiple elements. On the right is the degree of relevance of each element to the picture, which together form the label of this picture: $\bm{d}_{\text{image}}=\{0.03, 0.24, 0.21, 0.4, 0.03, 0.09\}$. Mountain and water tend to appear together, so the values of these two dimensions are both large. When cloud are present, the values describing the degree of the sky tend to be large. These are all manifestations of the correlations between the labels.} \label{fig:label_distribution}
\end{figure}
Whether in the MLL or LDL paradigm, the labels are not independent of each other. Therefore, making use of the correlations between labels is an important task during learning. In LDL, the labels with greater correlations tend to be assigned larger values simultaneously. For example, in the image classification task, the pair of labels ``seawater'' and ``beach'' tends to be highly correlated with the image. This situation is less likely to occur between labels with smaller correlations. For instance, in the movie classification task, labels like ``love'' and ``horror'', ``technology'' are basically unlikely to be assigned larger values simultaneously. A more concrete example is illustrated in Fig.~\ref{fig:label_distribution}. Making good use of the correlations between labels can bring improvements in both effectiveness and performance for the algorithm. Current work mainly utilizes the correlation of labels through two methods. One is external regularization such as statistical information~\citep{jia2018ldllc, ren2019ldlsf}, graph methods~\citep{jin2024gldl, lin2024ldlhvlc}, or other special regularization items~\citep{wu2024ldlda, jia2021ldllrr, XU2025incomplete}. This is also the most commonly used method. The other is latent space learning such as low-rank methods~\citep{ren2019ldllclr, tang2020le}. The former adds regularization terms to the loss function based on the label correlations to constrain the training of the model; the latter directly alters the label space and learns the mapping from the input space to the new label space, then projects the prediction back to the original label space. These methods predominantly focus on modeling correlations within the output label space, whereas few works concentrate on utilizing label correlations to enrich the input feature space.

Deep Forest~\citep{zhou2017deep} is a deep learning framework that does not rely on backpropagation. Its implementation draws on the key success factors of deep neural networks: layer-by-layer processing, in-model feature transformation, and sufficient model complexity. Each layer of its cascade structure is an ensemble of multiple forests. Through this structure, features are extracted at each layer and passed to the next layer for processing. Due to its strong feature representation ability and fewer hyperparameters compared to neural networks, DF has achieved success in many tasks such as image classification~\citep{boualleg2019dfimage, burmeister2025dfimage}. It has also developed many different variations. \citet{chen2021highorder} utilize high-dimensional feature interactions as the internal representation features, further enhancing its representational ability; \citet{UTKIN2020imprecise} focuses on improving its robustness on small datasets by incorporating imprecise probabilities to handle the uncertainty in class predictions. Additionally, assigning different weights to the trees in the forest~\citep{utkin2018siamese} has been used to adapt the framework for metric learning. Some studies have also provided theoretical proof of its performance, for example~\cite{lyu2022region,lyu2022depth}. However, its exploration in non-single-label learning tasks (such as MLL, LDL) is still at an early stage. A major reason for this is that its characteristic of lacking a loss function makes it unable to effectively utilize traditional methods to leverage the correlation between labels. \citet{yang2020multi} apply DF to the MLL task, using a feature reuse mechanism to ensure the stability of the training. \citet{Ilidio2024weak} conduct exploration in the task of weak label learning. Although they both utilize the correlations between labels to some extent, this process is implicit rather than explicit. 

This paper proposes the ERDF method, which explicitly utilizes the inherent correlation between labels through a feature enhancement mechanism. This mechanism is seamlessly integrated with the cascade structure of DF, taking full advantage of its layer-by-layer in-model feature extraction and transformation capabilities. At the same time, to guarantee the quality of the newly generated features and maintain the overall stability of the training phase, ERDF uses the feature reuse mechanism to reuse poor features. The main contributions of this paper can be summarized as follows: 

\begin{itemize}
\item 
    We propose a new and simple approach to exploring the correlations between labels during the learning process. Unlike most of previous methods, it focuses on the input feature space.
\item 
    We apply this new mechanism to DF, supplemented by the feature reuse mechanism, and conduct a successful exploration of DF in the field of LDL.

\item 
    Experiments on multiple public datasets demonstrate the superiority of our algorithm. The ablation experiments demonstrate the effectiveness of our mechanism.
\end{itemize}
The remainder of the paper is organized as follows. Section~\ref{ch:preliminary} introduces some preliminaries. Section~\ref{ch:method} detailes our ERDF method, including two designed mechanisms. Section~\ref{ch:experiment} reports the experimental results followed by the conclusion of the paper in Section~\ref{ch:conclusion}.

\section{Preliminary}
\label{ch:preliminary}


In label distribution learning, $\mathcal{X}$ $\subseteq$ $\mathcal{R}^d$ is a $d$-dimentional feature space and $\mathcal{Y} = \{y_1, y_2, \dots, y_c\}$ is the set of $c$ possible labels. For the $i$-th instance $x_i$ $\in$ $\mathcal{X}$, its corresponding label is a label distribution, denoted as $\bm{d_i}=\{d_i^1, d_i^2,\dots d_i^c\}$. Here $d_i^j$ represents the degree of correlation between $x_i$ and $y_j$. It should be noted that $\bm{d_i}$ must satisfy two constraints: 
\begin{itemize}
    \item \textbf{Non-negativity:} $d_i^j \ge 0, \quad \forall j \in \{1, \dots, c\}$.
    \item \textbf{Normalization:} $\sum_{j=1}^{c} d_i^j = 1$.
\end{itemize}
Fig.~\ref{fig:label_distribution} is an example sample. The learning task is to learn a mapping function $f: \mathcal{X} \to \mathcal{D}$, which optimizes the performance of unseen instances on some specific metrics. $\mathcal{X}$ here is the input space and $\mathcal{D}$ is the space of all possible label distributions.

Unlike the evaluation metrics for traditional machine learning tasks, such as accuracy for classification and mean square error for regression, metrics for label distribution learning are aimed at describing the relationship between two probability distributions. Widely-used metrics~\citep{geng2016label}, which are divided into two types, are listed in Table~\ref{tab:ldl_metrics}.

\begin{table}[h!]
\centering
\renewcommand{\arraystretch}{2} 
\caption{Six evaluation metrics for label distribution learning are divided into two types. 
The arrow indicates the direction for better performance: $\uparrow$ signifies that a higher value is better, and $\downarrow$ signifies that a lower value is better.}
\label{tab:ldl_metrics}
\begin{tabular}{lll}
\hline
\textbf{Type} & \textbf{Measure} & \textbf{Formula} \\
\hline
\multirow{4}{*}{Distance $\downarrow$} 
& Chebyshev & $Dis_1(\bm{d}, \hat{\bm{d}}) = \max_j |d_j - \hat{d}_j|$ \\
& Clark & $Dis_2(\bm{d}, \hat{\bm{d}}) = \sqrt{\sum_{j=1}^{c} \frac{(d_j - \hat{d}_j)^2}{(d_j + \hat{d}_j)^2}}$ \\
& Canberra & $Dis_3(\bm{d}, \hat{\bm{d}}) = \sum_{j=1}^{c} \frac{|d_j - \hat{d}_j|}{d_j + \hat{d}_j}$ \\
& Kullback-Leibler & $Dis_4(\bm{d}, \hat{\bm{d}}) = \sum_{j=1}^{c} d_j \ln \frac{d_j}{\hat{d}_j}$ \\
\hline
\multirow{2}{*}{Similarity $\uparrow$} 
& Cosine & $Sim_1(\bm{d}, \hat{\bm{d}}) = \frac{\sum_{j=1}^{c} d_j \hat{d}_j}{\sqrt{\sum_{j=1}^{c} d_j^2} \sqrt{\sum_{j=1}^{c} \hat{d}_j^2}}$ \\
& Intersection & $Sim_2(\bm{d}, \hat{\bm{d}}) = \sum_{j=1}^{c} \min(d_j, \hat{d}_j)$ \\
\hline
\end{tabular}
\end{table}

The first type is distance metrics. For this kind of metric, the larger the number, the greater the distance between the two distributions, and the less accurate the prediction result will be. The Chebyshev distance describes the maximum difference in all dimensions between two distributions, capturing the most extreme errors in the prediction. The Clark distance and Canberra distance are the weighted forms of the Euclidean distance and the Manhattan distance, respectively. Under the condition where the differences in a certain dimension are the same, they are more sensitive to the smaller values. The KL divergence measures the information loss that occurs when approximating one distribution with another. It should be noted that it is not symmetric. 

The second type is similarity metrics, and it is easy to understand that the larger the number, the better the result. Cosine similarity describes the consistency of the directions of two vectors in a multidimensional space. It does not consider the magnitude of the values in each dimension, but only focuses on whether the trends of the vectors are similar. The Intersection similarity describes the area of overlap between two distributions. Contrary to cosine similarity, it mainly focuses on the magnitude of the values in each dimension and is used to reflect the degree of consistency between the two distributions in terms of numerical values.

\section{The Proposed Method}
\label{ch:method}

\subsection{Feature enhancement exploiting label correlation}
\label{ch:fe}

This mechanism is specifically designed to enable DF to take advantage of the inherent correlations among the labels, thereby facilitating better performance in label distribution learning. Given the DF's powerful ability to extract and utilize new features layer by layer, we leverage the correlations among labels to enhance the features at each layer. This mechanism can be divided into two stages: the learning stage and the enhancement stage. The detailed pseudo-code of the algorithm is presented in Algorithm~\ref{alg:fe}.

During the learning phase, it can be divided into three steps. The first step is to extract the relevance. We calculate the Pearson correlation matrix of the label matrix to obtain the correlation relationships between the labels. The value in the $i$-th row and $j$-th column of the correlation matrix, which represents the correlation coefficient between the $i$-th label and the $j$-th one, is calculated according to Eq.~\eqref{eq:pearson}. $N$ is the total number of samples, $d_{s}^{i}$ is the description degree of the $i$-th label for the $s$-th sample, and $\bar{d}^i$ is the mean description degree for the $i$-th label over all samples. The resulting matrix $\bm{C} \in \mathbb{R}^{c \times c}$, where $c$ is the number of classes, serves as an explicit representation of the global label correlations. This matrix is like a social network diagram, revealing positive or negative correlations between those labels with positive or negative numbers ranging from $-1$ to $1$. 

\begin{equation}
\label{eq:pearson}
r_{ij} = \frac{\sum_{s=1}^{N} (d_{s}^{i} - \bar{d}^i)(d_{s}^j - \bar{d}^j)}{\sqrt{\sum_{s=1}^{N} (d_{s}^i - \bar{d}^i)^2 \sum_{s=1}^{N} (d_{s}^{j} - \bar{d}^j)^2}}
\end{equation}

The second step is to extract the patterns. The information in the correlation matrix is vast and may be redundant. We use principal component analysis (PCA) to extract $k$ of the most representative relationship patterns from it. These patterns constitute the matrix $\bm{V}$, which consists of $k$ eigenvectors as shown in Eq.~\eqref{eq:pca}. Each pattern vector is the same dimensions as the label distribution of each sample, corresponding to one dimension of the enhanced feature later. 

\begin{equation}
\bm{V} = \text{PCA}(\bm{C}, k) = [\bm{v}_1, \bm{v}_2, \dots, \bm{v}_k]
\label{eq:pca}
\end{equation}

The third step is to calculate the target relationship values and train the feature enhancers. For the $j$-th relationship pattern vector $\bm{v}_j$, we perform a dot product between the global label matrix $\bm{D}$ and $\bm{v}_j$ to obtain the target vector $\bm{s}_j$, as shown in Eq.~\eqref{eq:target_calc}, where $\bm{D} \in \mathbb{R}^{N \times c}$ is the label matrix of all training samples. The resulting vector $\bm{s}_j \in \mathbb{R}^{N \times 1}$ represents the degree of similarity between all samples and the $j$-th relationship pattern. Subsequently, we train the $j$-th feature enhancer to learn the mapping from the input features $\bm{X}$ to this target vector $\bm{s}_j$.

\begin{equation}
\label{eq:target_calc}
\bm{s}_j = \bm{D} \bm{v}_j, \quad j \in \{1, \dots, k\}
\end{equation}

During the subsequent enhancement stage, we utilize all $k$ feature enhancers to generate k-dimensional enhanced features for each sample. These enhanced features are then concatenated with the original features to ultimately obtain the final feature vector.

\begin{algorithm}[ht]
\caption{Feature enhancement exploiting label correlation}
\label{alg:fe}
\small
\begin{algorithmic}[1]
\Require Feature matrix $\bm{X} \in \mathbb{R}^{N \times D}$; 
        Label matrix $\bm{D} \in \mathbb{R}^{N \times c}$; 
        Number of relational features to generate $k$.
\Ensure A set of trained enhancers $\mathcal{E} = \{e_1, \dots, e_k\}$; 
        Enhanced relational feature matrix $\bm{E} \in \mathbb{R}^{N \times k}$.
\vspace{0.2em}
\Procedure{Fit}{$\bm{X}, \bm{D}, k$} \Comment{Learning Phase}
    \LineComment{Compute label correlation matrix according to Eq.~\eqref{eq:pearson}}
    \State $\bm{C} \gets \text{CalculateCorrelationMatrix}(\bm{D})$ 
    
    \State $\bm{V} \gets \text{PCA}(\bm{C}, k)$  \Comment{Extract $k$ relationship vectors according to Eq.~\eqref{eq:pca}}

    \State $\mathcal{E} \gets \emptyset$ \Comment{Initialize empty enhancer set}
    \For{$j = 1$ to $k$}
        \State $\bm{s}_j \gets \bm{D} \bm{v}_j$ \Comment{Ideal scores calculated according to Eq.~\eqref{eq:target_calc}}; 
        \State $e_j \gets \text{RandomForestRegressor}(\bm{X}, \bm{s}_j)$; $\mathcal{E} \gets \mathcal{E} \cup \{e_j\}$
    \EndFor
    \State \Return $\mathcal{E}$
\EndProcedure

\Statex
\vspace{0.2em}
\Procedure{Transform}{$X, \mathcal{E}$} \Comment{Enhancement Phase}
    \State $k \gets |\mathcal{E}|$; $\bm{E} \gets \bm{0}^{N \times k}$ 
    \For{$j = 1$ to $k$}
        \State $\hat{\bm{s}}_j \gets e_j.\text{predict}(\bm{X})$
        \State $\text{SetColumn}(\bm{E}, j, \hat{\bm{s}}_j)$ \Comment{Set the j-th column}
    \EndFor
    \State \Return $\bm{E}$
\EndProcedure
\end{algorithmic}
\end{algorithm}

Let us consider the movie classification task mentioned earlier. Assume that the label set $\mathcal{Y} = \{\text{action}, \allowbreak \text{comedy}, \allowbreak \text{romance}, \allowbreak \text{horror}, \allowbreak \text{science fiction}\}$, We first calculate the correlation matrix. Obviously, action and horror are likely to be positively correlated, while comedy and horror are likely to be negatively correlated. Through principal component analysis, we may obtain two distinct relationship patterns. The first relationship pattern represents the meaning of ``blockbuster''; in this pattern, the values of action and science fiction are relatively large. The second relationship pattern represents the meaning of ``youth''; in this pattern, the value of romance is relatively large. Subsequently, for each relationship pattern, we calculate the similarity between the label distribution of all samples and the target relationship pattern, and then learn an enhancer to map from the original feature vector to the relationship pattern values, which is used to enhance the original features.

DF can extract new feature vectors to perform in-model feature transform layer by layer. In this mechanism, new features are enhanced at each layer, which is perfectly adaptive to the layer-by-layer feature transformation process in DF. This is a key factor for the success of this mechanism. It should be noted that the reason for training the feature enhancer instead of directly using the calculated relationship values as new features is that during generalization, the true label distribution of the samples is invisible.

\subsection{Measure-aware feature reuse}
\label{ch:fr}

Measure-aware feature reuse~\citep{yang2020multi} is a mechanism that ensures the quality of features extracted by the cascade structure of DF and promotes the stability of layer-by-layer training. ERDF not only extracts new features like standard  DF, but also enhances the features through the enhancers layer by layer. This results in more new features at each layer, which leads to a greater need for the feature reuse mechanism to ensure the stability of training and to perform reuse operations for new features of low quality.

After the training on the $l$-th layer is completed ($l>1$), we will obtain the validation result matrix $\bm{H}_l$ for all samples. $\bm{H}_l$ is viewed as the first part of the new features, and we will concatenate it with the original feature, just as DF does. Then, ERDF will use the feature enhancement mechanism described in Section~\ref{ch:fe} to enhance the combined features, obtaining the second part of the new features $\bm{E}_l$. By concatenating $\bm{H}_l$ and $\bm{E}_l$ according to Eq.~\eqref{eq:feat_concat}, we get all the new features $\bm{F}_{new}^{(l)}$ for this layer. At this time, we cannot guarantee that all the features in $\bm{F}_{new}^{(l)}$ are of high quality and beneficial for the subsequent training. If we directly pass it to the subsequent layers, it may cause instability in the training process. The noise in the new features will gradually expand as the layer deepens, causing a disastrous impact on the training.

\begin{equation}
\label{eq:feat_concat}
\mathbf{F}_{\text{new}}^{(l)} = [\mathbf{H}_l, \mathbf{E}_l]
\end{equation}

The measure-aware feature reuse mechanism comes into play in this scenario. First, based on actual requirements, we can select a measurement metric $\mathcal{M}$. By comparing the prediction results of this layer ($\bm{H}_l$), with those of the previous layer ($\bm{H}_{l-1}$), we can quickly identify the sample set $S$ that has deteriorated in performance on metric $\mathcal{M}$ after the training of the $l$-th layer. Subsequently, using a certain threshold $\tau$ as the boundary, we determine the specific subset of samples $S_r$ requiring reuse through Eq.~\eqref{eq:Sr}:

\begin{equation}
\label{eq:Sr}
S_r = \left\{ s \in \{1, \dots, N\} \;\middle|\; \mathbb{I}(m_{l,s} > m_{l-1,s}) \cdot \mathbb{I}(m_{l,s} > \tau) = 1 \right\}
\end{equation}
where $\mathbb{I}(\cdot)$ is the indicator function. $m_{l,s} = \mathcal{M}(\bm{d}_s, \bm{h}_{l,s})$ denotes the calculated metric value obtained by comparing the validation prediction $\mathbf{h}_{l,s}$ of the $s$-th sample at layer $l$ with its corresponding ground-truth label distribution $\bm{d}_s$. It is worth noting that the inequalities in Eq.~\eqref{eq:Sr} assume that $\mathcal{M}$ is a distance metric (e.g., KL divergence) where a lower value indicates better performance. If a similarity metric (e.g., Cosine similarity) is used, where a higher value is better, the direction of these inequalities should be reversed.

ERDF then applies the reuse operation for the new features of samples in $S_r$, which are obtained through this layer's training. Concretely, ERDF replaces the rows of $\bm{F}_{new}^{(l)}$ which represent the samples in $S_r$, with the corresponding rows in matrix $\bm{G}_{l-1}$ which are the final new features of these samples in the previous layer. Now, new features in $\bm{F}_{new}^{(l)}$ are more reliable, and we record the newly obtained matrix as $\bm{G}_l$, which are the final new features of all samples after this layer's training. Then, we concatenate $\bm{G}_l$ with the original features and pass it to the next layer for training. 

As for the selection of the threshold $\tau$, we simply use the mean score of all the samples in $S$ which is calculated on $\mathcal{M}$ as the boundary for elimination. We opt for partial rather than complete elimination because the latter is an overly greedy approach, assuming current status guarantees the best. In practice, a deterioration in a specific metric does not always imply failure. Novel feature perspectives may induce temporary fluctuations which is sometimes a normal part of the learning dynamics.
\subsection{The framework}


\begin{figure}[h]
\centering
\includegraphics[width=\textwidth]{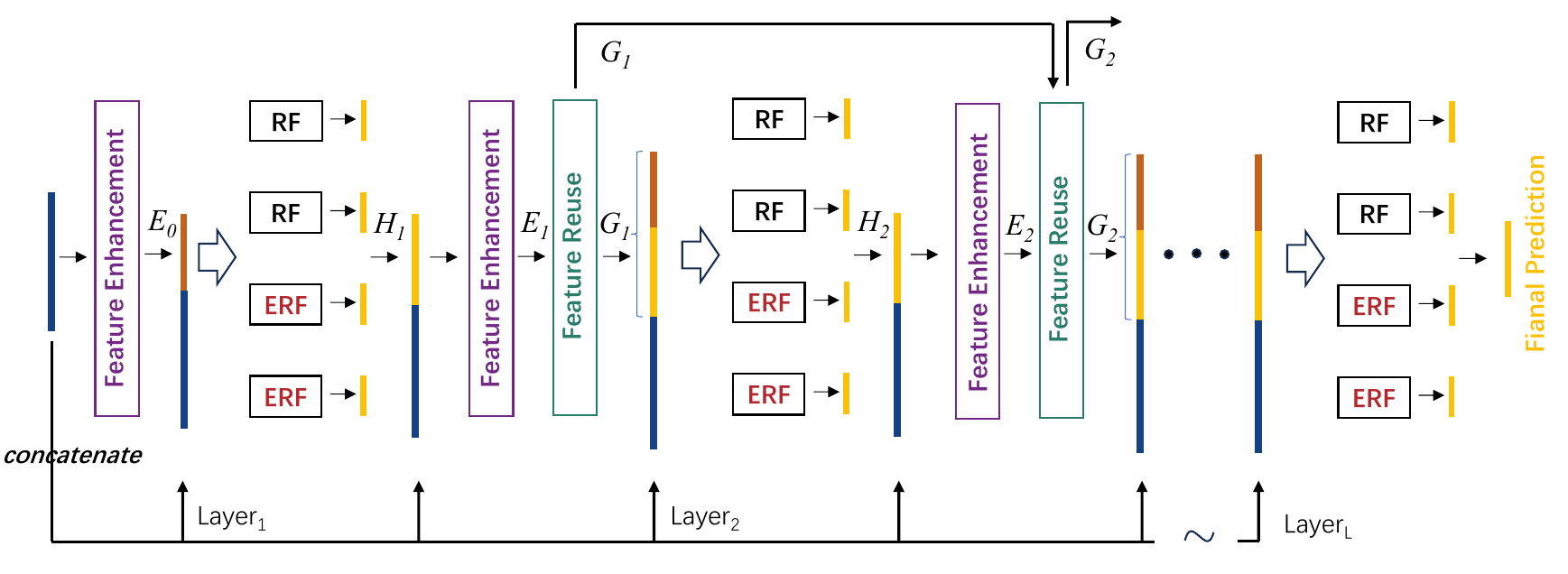}
\caption{This figure clearly presents the overall framework of ERDF. Each layer is composed of RF and ERF. After the training of the $l$-th layer, the first part of the new features $\bm{H}_l$ is obtained. These new features are composed of the validation prediction results of all forests in this layer. Then, the feature enhancement mechanism is conducted to obtain the second part of the new features $\bm{E}_l$. These two parts are concatenated to form $\bm{F}_{new}^{(l)}$, which is input into the feature reuse module. After reuse according to $\bm{G}_{l-1}$, the final new features $\bm{G}_l$ of this layer are obtained and concatenated with the original features before being passed to the next layer.} \label{fig:framework}
\end{figure}

The overall framework of ERDF is shown in Fig.~\ref{fig:framework}. It mainly adds a feature enhancement module (described in Section~\ref{ch:fe}) and a feature reuse module (described in Section~\ref{ch:fr}) on the basis of traditional DF. 

Before starting training, an initial feature enhancement is conducted, and the obtained enhancer $\mathcal{E}_0$ is stored. Then, the feature vector is input into the cascaded forest for training. Each layer contains two types of forests, random forest (RF) and extremely random forest (ERF), to ensure diversity. 

After the training of the $l$-th layer, the validation prediction result $\bm{H}_l$ of all samples is given as the first part of the new feature, and it is also used to evaluate the current layer's training performance. Then $\bm{H}_l$ is concatenated with the original feature, input into the feature enhancement module for enhancement. The second part of the new feature $\bm{E}_l$ and the current layer's enhancers $\mathcal{E}_l$ are obtained. The two parts of the new features are concatenated to form $\bm{F}_{new}^{(l)}$, which is input into the feature reuse module for reuse. After the reuse operation is finished, we can get this layer's final new feature matrix $\bm{G}_l$ and the elimination threshold $\tau_l$ of this layer. 

Note that the feature reuse module of the first layer cannot perform a reuse operation due to the lack of former information, and will directly record $F_{new}^{(0)}$ as the final new feature matrix $\bm{G}_0$ for subsequent use. $\bm{G}_l$ will be concatenated with the original feature, passed to the next layer for training, and also passed to the feature reuse module of the next layer for judgment on whether the performance has declined. The number of layers of the cascaded forest is self-adapting. Training will stop automatically when there is no significant improvement in performance. It should be noted that we place the feature reuse mechanism after the feature enhancement mechanism. This fixed order cannot be changed. If reuse is performed before enhancement, only the first part of the new features $\bm{H}_l$ will be reused, and it cannot ensure that the quality of the second part of the new features $\bm{E}_i$ remains stable.

For unseen samples, the initialization enhancement is first performed with $\mathcal{E}_0$, and then input into the cascade for prediction. Similar to the training process, after obtaining the validation prediction result $\bm{H}_l$ of the $l$-th layer, it will be processed by two modules. The difference is that in the feature enhancement module, the trained enhancer is directly used for enhancement, and in the feature reuse module, the threshold stored during the training stage is directly used to screen the reused samples. When the layer with the highest validation performance during the training process gives the prediction, the transmission is terminated, and it is viewed as the final prediction.

\section{Experiments}
\label{ch:experiment}

In this section, we conduct experiments on our ERDF algorithm on multiple public LDL datasets to validate its performance. We compare it with several commonly used LDL algorithms and demonstrate the superiority of ERDF. We also conducte ablation experiments on the two mechanisms of ERDF to demonstrate their necessity. 

\subsection{Dataset and configuration}

 Benchmark and abalation experiments are conducted on five commonly used public datasets. Table~\ref{tab:datasets} shows the detailed information with sample sizes ranging from 1500 to 7755, feature numbers ranging from 168 to 1869, and the number of classes (i.e., the dimension of label distribution) ranging from 5 to 9. In the experiments, we randomly divide the datasets into training sets and test sets in the ratio of 8:2. To reduce the randomness of the experiments, for each dataset, we split the samples using three different random seeds, and finally take the mean and standard deviation of the results as the final reported result. 

\begin{table}[ht]
\centering
\begin{tabular}{lccc}
\hline
Dataset & Samples & Features & Classes\\
\hline
Movie & 7755 & 1869 & 5 \\
Natural\_Scene & 2000 & 294 & 9 \\
SBU\_3DFE & 2500 & 243 & 6 \\
emotion6 & 1980 & 168 & 7 \\
SCUT\_FBP & 1500 & 300 & 5 \\
\hline
\end{tabular}
\caption{Information about the five datasets we used in our experiments.}\label{tab:datasets}
\end{table}

\begin{table}[ht]
\centering
\small
\resizebox{\textwidth}{!}{
\begin{tabular}{l l c c c c c}
\toprule
Dataset & Metric & AA-KNN & AA-BP & StructRF & LDL-SCL & ERDF(ours) \\
\midrule
\multirow{6}{*}{Movie} & KL div $\downarrow$ & 0.1133 $\pm$ 0.0010 $^{\bullet}$ & 0.1318 $\pm$ 0.0021 $^{\bullet}$ & 0.0926 $\pm$ 0.0025 $^{\bullet}$ & 0.1024 $\pm$ 0.0024 $^{\bullet}$ & \textbf{0.0600 $\pm$ 0.0014} \\
 & Chebyshev $\downarrow$ & 0.1227 $\pm$ 0.0004 $^{\bullet}$ & 0.1363 $\pm$ 0.0011 $^{\bullet}$ & 0.1108 $\pm$ 0.0010 $^{\bullet}$ & 0.1189 $\pm$ 0.0014 $^{\bullet}$ & \textbf{0.0939 $\pm$ 0.0011} \\
 & Clark $\downarrow$ & 0.5463 $\pm$ 0.0037 $^{\bullet}$ & 0.6146 $\pm$ 0.0053 $^{\bullet}$ & 0.5035 $\pm$ 0.0039 $^{\bullet}$ & 0.5309 $\pm$ 0.0105 $^{\bullet}$ & \textbf{0.4289 $\pm$ 0.0070} \\
 & Canberra $\downarrow$ & 1.0462 $\pm$ 0.0050 $^{\bullet}$ & 1.1636 $\pm$ 0.0089 $^{\bullet}$ & 0.9597 $\pm$ 0.0069 $^{\bullet}$ & 1.0143 $\pm$ 0.0170 $^{\bullet}$ & \textbf{0.8026 $\pm$ 0.0118} \\
 & Cosine $\uparrow$ & 0.9251 $\pm$ 0.0006 $^{\bullet}$ & 0.9144 $\pm$ 0.0006 $^{\bullet}$ & 0.9391 $\pm$ 0.0017 $^{\bullet}$ & 0.9321 $\pm$ 0.0011 $^{\bullet}$ & \textbf{0.9604 $\pm$ 0.0006} \\
 & Intersection $\uparrow$ & 0.8246 $\pm$ 0.0004 $^{\bullet}$ & 0.8051 $\pm$ 0.0010 $^{\bullet}$ & 0.8424 $\pm$ 0.0015 $^{\bullet}$ & 0.8317 $\pm$ 0.0020 $^{\bullet}$ & \textbf{0.8715 $\pm$ 0.0013} \\
\midrule
\multirow{6}{*}{Natural} & KL div $\downarrow$ & 1.0357 $\pm$ 0.0575 $^{\bullet}$ & 1.2452 $\pm$ 0.0588 $^{\bullet}$ & 0.6689 $\pm$ 0.0295 $^{\bullet}$ & 0.8283 $\pm$ 0.0118 $^{\bullet}$ & \textbf{0.5216 $\pm$ 0.0155} \\
 & Chebyshev $\downarrow$ & 0.3134 $\pm$ 0.0100 $^{\bullet}$ & 0.3922 $\pm$ 0.0124 $^{\bullet}$ & 0.2794 $\pm$ 0.0172 $^{\bullet}$ & 0.3329 $\pm$ 0.0083 $^{\bullet}$ & \textbf{0.2221 $\pm$ 0.0049} \\
 & Clark $\downarrow$ & \textbf{1.9085 $\pm$ 0.0214} $^{\circ}$ & 2.5099 $\pm$ 0.0082 $^{\bullet}$ & 2.3465 $\pm$ 0.0175 & 2.4682 $\pm$ 0.0116 $^{\bullet}$ & 2.3595 $\pm$ 0.0209 \\
 & Canberra $\downarrow$ & \textbf{4.5686 $\pm$ 0.0701} $^{\circ}$ & 7.0532 $\pm$ 0.0061 $^{\bullet}$ & 6.2559 $\pm$ 0.0761 $^{\circ}$ & 6.7833 $\pm$ 0.0468 $^{\bullet}$ & 6.3066 $\pm$ 0.0777 \\
 & Cosine $\uparrow$ & 0.7150 $\pm$ 0.0022 $^{\bullet}$ & 0.5582 $\pm$ 0.0163 $^{\bullet}$ & 0.7820 $\pm$ 0.0089 $^{\bullet}$ & 0.7284 $\pm$ 0.0036 $^{\bullet}$ & \textbf{0.8343 $\pm$ 0.0040} \\
 & Intersection $\uparrow$ & 0.5621 $\pm$ 0.0058 $^{\bullet}$ & 0.3665 $\pm$ 0.0052 $^{\bullet}$ & 0.5927 $\pm$ 0.0151 $^{\bullet}$ & 0.4994 $\pm$ 0.0047 $^{\bullet}$ & \textbf{0.6704 $\pm$ 0.0062} \\
\midrule
\multirow{6}{*}{SBU} & KL div $\downarrow$ & 0.0802 $\pm$ 0.0011 $^{\bullet}$ & 0.1065 $\pm$ 0.0167 $^{\bullet}$ & 0.0573 $\pm$ 0.0004 $^{\bullet}$ & 0.0634 $\pm$ 0.0005 $^{\bullet}$ & \textbf{0.0414 $\pm$ 0.0005} \\
 & Chebyshev $\downarrow$ & 0.1273 $\pm$ 0.0022 $^{\bullet}$ & 0.1378 $\pm$ 0.0018 $^{\bullet}$ & 0.1104 $\pm$ 0.0013 $^{\bullet}$ & 0.1199 $\pm$ 0.0015 $^{\bullet}$ & \textbf{0.0865 $\pm$ 0.0013} \\
 & Clark $\downarrow$ & 0.4012 $\pm$ 0.0017 $^{\bullet}$ & 0.5113 $\pm$ 0.0668 $^{\bullet}$ & 0.3444 $\pm$ 0.0009 $^{\bullet}$ & 0.3695 $\pm$ 0.0017 $^{\bullet}$ & \textbf{0.2746 $\pm$ 0.0020} \\
 & Canberra $\downarrow$ & 0.8309 $\pm$ 0.0039 $^{\bullet}$ & 1.0997 $\pm$ 0.1475 $^{\bullet}$ & 0.7276 $\pm$ 0.0060 $^{\bullet}$ & 0.7960 $\pm$ 0.0043 $^{\bullet}$ & \textbf{0.5772 $\pm$ 0.0044} \\
 & Cosine $\uparrow$ & 0.9219 $\pm$ 0.0011 $^{\bullet}$ & 0.9039 $\pm$ 0.0103 $^{\bullet}$ & 0.9432 $\pm$ 0.0006 $^{\bullet}$ & 0.9376 $\pm$ 0.0006 $^{\bullet}$ & \textbf{0.9587 $\pm$ 0.0006} \\
 & Intersection $\uparrow$ & 0.8487 $\pm$ 0.0010 $^{\bullet}$ & 0.8106 $\pm$ 0.0196 $^{\bullet}$ & 0.8691 $\pm$ 0.0007 $^{\bullet}$ & 0.8575 $\pm$ 0.0007 $^{\bullet}$ & \textbf{0.8968 $\pm$ 0.0007} \\
\midrule
\multirow{6}{*}{SCUT} & KL div $\downarrow$ & 0.5241 $\pm$ 0.0049 $^{\bullet}$ & 0.3851 $\pm$ 0.0140 $^{\bullet}$ & 0.3337 $\pm$ 0.0094 $^{\bullet}$ & 0.5647 $\pm$ 0.0056 $^{\bullet}$ & \textbf{0.2949 $\pm$ 0.0134} \\
 & Chebyshev $\downarrow$ & 0.2537 $\pm$ 0.0035 $^{\bullet}$ & 0.2629 $\pm$ 0.0051 $^{\bullet}$ & 0.2303 $\pm$ 0.0024 $^{\bullet}$ & 0.3418 $\pm$ 0.0073 $^{\bullet}$ & \textbf{0.2092 $\pm$ 0.0049} \\
 & Clark $\downarrow$ & \textbf{1.3043 $\pm$ 0.0116} $^{\circ}$ & 1.3846 $\pm$ 0.0078 $^{\bullet}$ & 1.3675 $\pm$ 0.0032 $^{\bullet}$ & 1.4561 $\pm$ 0.0050 $^{\bullet}$ & 1.3523 $\pm$ 0.0027 \\
 & Canberra $\downarrow$ & \textbf{2.3985 $\pm$ 0.0411} & 2.5792 $\pm$ 0.0194 $^{\bullet}$ & 2.5220 $\pm$ 0.0037 $^{\bullet}$ & 2.8076 $\pm$ 0.0234 $^{\bullet}$ & 2.4651 $\pm$ 0.0122 \\
 & Cosine $\uparrow$ & 0.8294 $\pm$ 0.0045 $^{\bullet}$ & 0.8414 $\pm$ 0.0066 $^{\bullet}$ & 0.8624 $\pm$ 0.0045 $^{\bullet}$ & 0.7598 $\pm$ 0.0053 $^{\bullet}$ & \textbf{0.8782 $\pm$ 0.0063} \\
 & Intersection $\uparrow$ & 0.6942 $\pm$ 0.0058 $^{\bullet}$ & 0.6792 $\pm$ 0.0074 $^{\bullet}$ & 0.7210 $\pm$ 0.0025 $^{\bullet}$ & 0.5750 $\pm$ 0.0064 $^{\bullet}$ & \textbf{0.7552 $\pm$ 0.0054} \\
\midrule
\multirow{6}{*}{emotion6} & KL div $\downarrow$ & 0.9038 $\pm$ 0.0264 $^{\bullet}$ & 0.6802 $\pm$ 0.0031 $^{\bullet}$ & 0.5893 $\pm$ 0.0093 $^{\bullet}$ & 0.5862 $\pm$ 0.0230 $^{\bullet}$ & \textbf{0.5111 $\pm$ 0.0252} \\
 & Chebyshev $\downarrow$ & 0.3326 $\pm$ 0.0113 $^{\bullet}$ & 0.3413 $\pm$ 0.0033 $^{\bullet}$ & 0.3137 $\pm$ 0.0051 $^{\bullet}$ & 0.3108 $\pm$ 0.0090 $^{\bullet}$ & \textbf{0.2869 $\pm$ 0.0056} \\
 & Clark $\downarrow$ & 1.7087 $\pm$ 0.0146 $^{\bullet}$ & 1.6686 $\pm$ 0.0181 & 1.6569 $\pm$ 0.0210 $^{\bullet}$ & 1.6503 $\pm$ 0.0221 & \textbf{1.6495 $\pm$ 0.0230} \\
 & Canberra $\downarrow$ & 3.8728 $\pm$ 0.0475 $^{\bullet}$ & 3.7883 $\pm$ 0.0280 & 3.7285 $\pm$ 0.0502 & 3.7000 $\pm$ 0.0510 & \textbf{3.6858 $\pm$ 0.0684} \\
 & Cosine $\uparrow$ & 0.6561 $\pm$ 0.0120 $^{\bullet}$ & 0.6657 $\pm$ 0.0026 $^{\bullet}$ & 0.7125 $\pm$ 0.0026 $^{\bullet}$ & 0.7152 $\pm$ 0.0079 $^{\bullet}$ & \textbf{0.7496 $\pm$ 0.0078} \\
 & Intersection $\uparrow$ & 0.5534 $\pm$ 0.0079 $^{\bullet}$ & 0.5440 $\pm$ 0.0014 $^{\bullet}$ & 0.5793 $\pm$ 0.0042 $^{\bullet}$ & 0.5833 $\pm$ 0.0072 $^{\bullet}$ & \textbf{0.6238 $\pm$ 0.0089} \\
\midrule
\multicolumn{2}{l}{Avg. Rank} & 3.60 & 4.60 & 2.27 & 3.33 & 1.20 \\
\bottomrule
\end{tabular}
}
\caption{This table presents the results of ERDF and four common LDL algorithms on five datasets. All results are reported as mean $\pm$ standard deviation over three random splits. The arrow beside the metric indicates the direction for better performance. The optimal value of each row is displayed in bold. we use • (or ◦) to indicate that ERDF is significantly better (or worse) than the corresponding method.} \label{tab:baseline}

\end{table}

We use the six metrics shown in Table~\ref{tab:ldl_metrics} to conduct a comprehensive evaluation of the model from both the distance and similarity perspectives. We compare ERDF with four commonly used LDL algorithms. They are AA-KNN~\citep{geng2016label}, SA-BFGS~\citep{geng2016label}, StructRF~\citep{chen2018structrf}, and LDL-SCL~\citep{jia2019ldlscl}, respectively. AA-KNN predicts the label distribution of a new sample by finding its $k$ nearest neighbors in the feature space and averaging their label distributions. SA-BFGS constructs the learning process as an optimization problem and uses the BFGS optimizer to efficiently solve the model parameters. StructRF extends the traditional random forest framework to the LDL task. It treats the label distribution of each sample as a whole, enabling the model to directly predict the complete label distribution at the leaf nodes instead of making separate predictions for each dimension. LDL-SCL is an advanced algorithm that utilizes subspace clustering to leverage label correlations. It first clusters instances based on their labels and then enhances the original features with a local correlation vector representing the influence of these clusters.

Hyperparameters of ERDF are set as follows. We use a decision tree similar to StructTree~\citep{chen2018structrf} as the base learner which considers the label distribution of each sample as a whole. For each unseen sample that falls under a certain leaf node, the mean of the label distributions of all samples at that leaf node is taken as the final prediction result. The difference is that we use the KL divergence as the splitting criterion. The maximum depth of each tree is set to 10, and the minimum sample number of leaf nodes is set to 2. Each forest contains 100 decision trees, and each layer of the cascade forest consists of two random forests and two completely random forests. The maximum number of layers in the cascade forest is 10, and the early stopping tolerance is set to 1, which means that the training will be terminated in advance if there is no performance improvement for two consecutive layers before reaching the maximum number of layers. In the feature reuse mechanism, we use the KL divergence for performance measurement. In the feature enhancement mechanism, $k$ is set to 5, which means extracting five relationship patterns from the relationship matrix to generate 5-dimensional enhanced features for each instance.

The parameters of the comparison algorithms are listed as follows: In AA-KNN, $k$ is set to 5 and the minkowski distance metric is used to find the $k$ nearest neighbor of an instance. The maximum number of iterations for SA-BFGS is set to 500. In StructRF, the number of trees and the maximum depth are 100 and 10, respectively, which are consistent with our ERDF. In LDL-SCL, for each dataset, $\lambda_{1}$, $\lambda_{2}$, and $\lambda_{3}$ are tuned from $\{10^{-1}, 10^{-2}, 10^{-3}\}$, and the number of clusters $c$ is tuned in the set $\{5, 10\}$.

\subsection{Benchmark experiment}

As shown in Table~\ref{tab:baseline}, our proposed ERDF achieves state-of-the-art performance with a top average rank of 1.20. The foundational LDL algorithms, AA-KNN and SA-BFGS, which do not explicitly model label correlations, rank the lowest. This outcome empirically validates the central premise that leveraging label correlations is critical for success in LDL tasks.
StructRF performs competitively (average rank 2.27) by implicitly capturing correlations among labels. Its efficacy can be attributed to the non-linear modeling power of tree-based ensembles. Nevertheless, its performance is surpassed by ERDF, which benefits from a more potent and explicit mechanism for exploiting label correlations within a more powerful ensemble architecture. An insightful comparison can be drawn with LDL-SCL. Similar to ERDF, it explicitly enhances features using label correlations. However, LDL-SCL exhibits high sensitivity to its hyperparameters, requiring dataset-specific tuning to reach its optimal performance. In contrast, ERDF demonstrates remarkable robustness, achieving superior results with a consistent set of hyperparameters across all datasets, which highlights the stability and practical advantages of our proposed architecture.

\begin{table}[ht!]
\centering
\small
\resizebox{\textwidth}{!}{
\begin{tabular}{l l c c c c}
\toprule
Dataset & Metric & w/o fe \& fr & w/o fe & w/o fr & ERDF (ours) \\
\midrule
\multirow{6}{*}{Movie} & KL div $\downarrow$ & 0.0921 $\pm$ 0.0018 $^{\bullet}$ & 0.0698 $\pm$ 0.0011 $^{\bullet}$ & 0.0898 $\pm$ 0.0034 $^{\bullet}$ & \textbf{0.0600 $\pm$ 0.0014} \\
 & Chebyshev $\downarrow$ & 0.1120 $\pm$ 0.0010 $^{\bullet}$ & 0.1018 $\pm$ 0.0007 $^{\bullet}$ & 0.1082 $\pm$ 0.0020 $^{\bullet}$ & \textbf{0.0939 $\pm$ 0.0011} \\
 & Clark $\downarrow$ & 0.5129 $\pm$ 0.0070 $^{\bullet}$ & 0.4640 $\pm$ 0.0053 $^{\bullet}$ & 0.4917 $\pm$ 0.0074 $^{\bullet}$ & \textbf{0.4289 $\pm$ 0.0070} \\
 & Canberra $\downarrow$ & 0.9745 $\pm$ 0.0111 $^{\bullet}$ & 0.8712 $\pm$ 0.0090 $^{\bullet}$ & 0.9378 $\pm$ 0.0144 $^{\bullet}$ & \textbf{0.8026 $\pm$ 0.0118} \\
 & Cosine $\uparrow$ & 0.9395 $\pm$ 0.0009 $^{\bullet}$ & 0.9537 $\pm$ 0.0002 $^{\bullet}$ & 0.9408 $\pm$ 0.0024 $^{\bullet}$ & \textbf{0.9604 $\pm$ 0.0006} \\
 & Intersection $\uparrow$ & 0.8406 $\pm$ 0.0011 $^{\bullet}$ & 0.8594 $\pm$ 0.0006 $^{\bullet}$ & 0.8462 $\pm$ 0.0030 $^{\bullet}$ & \textbf{0.8715 $\pm$ 0.0013} \\
\midrule
\multirow{6}{*}{Natural} & KL div $\downarrow$ & 0.5912 $\pm$ 0.0139 $^{\bullet}$ & 0.5617 $\pm$ 0.0139 $^{\bullet}$ & 0.5866 $\pm$ 0.0150 $^{\bullet}$ & \textbf{0.5216 $\pm$ 0.0155} \\
 & Chebyshev $\downarrow$ & 0.2535 $\pm$ 0.0079 $^{\bullet}$ & 0.2436 $\pm$ 0.0074 $^{\bullet}$ & 0.2365 $\pm$ 0.0040 $^{\bullet}$ & \textbf{0.2221 $\pm$ 0.0049} \\
 & Clark $\downarrow$ & 2.4071 $\pm$ 0.0143 $^{\bullet}$ & 2.4000 $\pm$ 0.0169 $^{\bullet}$ & 2.3696 $\pm$ 0.0161 & \textbf{2.3595 $\pm$ 0.0209} \\
 & Canberra $\downarrow$ & 6.4948 $\pm$ 0.0622 $^{\bullet}$ & 6.4612 $\pm$ 0.0705 $^{\bullet}$ & 6.3594 $\pm$ 0.0587 & \textbf{6.3066 $\pm$ 0.0777} \\
 & Cosine $\uparrow$ & 0.8139 $\pm$ 0.0036 $^{\bullet}$ & 0.8247 $\pm$ 0.0038 $^{\bullet}$ & 0.8139 $\pm$ 0.0044 $^{\bullet}$ & \textbf{0.8343 $\pm$ 0.0040} \\
 & Intersection $\uparrow$ & 0.6194 $\pm$ 0.0074 $^{\bullet}$ & 0.6332 $\pm$ 0.0073 $^{\bullet}$ & 0.6561 $\pm$ 0.0053 $^{\bullet}$ & \textbf{0.6704 $\pm$ 0.0062} \\
\midrule
\multirow{6}{*}{SBU} & KL div $\downarrow$ & 0.0546 $\pm$ 0.0000 $^{\bullet}$ & 0.0477 $\pm$ 0.0004 $^{\bullet}$ & 0.0581 $\pm$ 0.0014 $^{\bullet}$ & \textbf{0.0414 $\pm$ 0.0005} \\
 & Chebyshev $\downarrow$ & 0.1083 $\pm$ 0.0013 $^{\bullet}$ & 0.1008 $\pm$ 0.0014 $^{\bullet}$ & 0.1019 $\pm$ 0.0004 $^{\bullet}$ & \textbf{0.0865 $\pm$ 0.0013} \\
 & Clark $\downarrow$ & 0.3343 $\pm$ 0.0004 $^{\bullet}$ & 0.3123 $\pm$ 0.0002 $^{\bullet}$ & 0.3138 $\pm$ 0.0007 $^{\bullet}$ & \textbf{0.2746 $\pm$ 0.0020} \\
 & Canberra $\downarrow$ & 0.7149 $\pm$ 0.0021 $^{\bullet}$ & 0.6665 $\pm$ 0.0019 $^{\bullet}$ & 0.6482 $\pm$ 0.0026 $^{\bullet}$ & \textbf{0.5772 $\pm$ 0.0044} \\
 & Cosine $\uparrow$ & 0.9459 $\pm$ 0.0001 $^{\bullet}$ & 0.9525 $\pm$ 0.0005 $^{\bullet}$ & 0.9414 $\pm$ 0.0013 $^{\bullet}$ & \textbf{0.9587 $\pm$ 0.0006} \\
 & Intersection $\uparrow$ & 0.8719 $\pm$ 0.0001 $^{\bullet}$ & 0.8807 $\pm$ 0.0003 $^{\bullet}$ & 0.8810 $\pm$ 0.0002 $^{\bullet}$ & \textbf{0.8968 $\pm$ 0.0007} \\
\midrule
\multirow{6}{*}{SCUT} & KL div $\downarrow$ & 0.3474 $\pm$ 0.0141 $^{\bullet}$ & 0.3349 $\pm$ 0.0141 $^{\bullet}$ & 0.4350 $\pm$ 0.0321 $^{\bullet}$ & \textbf{0.2949 $\pm$ 0.0134} \\
 & Chebyshev $\downarrow$ & 0.2401 $\pm$ 0.0048 $^{\bullet}$ & 0.2382 $\pm$ 0.0044 $^{\bullet}$ & 0.2455 $\pm$ 0.0098 $^{\bullet}$ & \textbf{0.2092 $\pm$ 0.0049} \\
 & Clark $\downarrow$ & 1.3681 $\pm$ 0.0034 $^{\bullet}$ & 1.3639 $\pm$ 0.0040 & 1.4051 $\pm$ 0.0128 $^{\bullet}$ & \textbf{1.3523 $\pm$ 0.0027} \\
 & Canberra $\downarrow$ & 2.5261 $\pm$ 0.0123 $^{\bullet}$ & 2.5128 $\pm$ 0.0136 & 2.6361 $\pm$ 0.0446 $^{\bullet}$ & \textbf{2.4651 $\pm$ 0.0122} \\
 & Cosine $\uparrow$ & 0.8548 $\pm$ 0.0064 $^{\bullet}$ & 0.8590 $\pm$ 0.0063 $^{\bullet}$ & 0.8263 $\pm$ 0.0140 $^{\bullet}$ & \textbf{0.8782 $\pm$ 0.0063} \\
 & Intersection $\uparrow$ & 0.7095 $\pm$ 0.0051 $^{\bullet}$ & 0.7135 $\pm$ 0.0048 $^{\bullet}$ & 0.7065 $\pm$ 0.0122 $^{\bullet}$ & \textbf{0.7552 $\pm$ 0.0054} \\
\midrule
\multirow{6}{*}{emotion6} & KL div $\downarrow$ & 0.5700 $\pm$ 0.0205 $^{\bullet}$ & 0.5338 $\pm$ 0.0173 & 0.7410 $\pm$ 0.0335 $^{\bullet}$ & \textbf{0.5111 $\pm$ 0.0252} \\
 & Chebyshev $\downarrow$ & 0.3121 $\pm$ 0.0074 $^{\bullet}$ & 0.3011 $\pm$ 0.0065 $^{\bullet}$ & 0.3355 $\pm$ 0.0029 $^{\bullet}$ & \textbf{0.2869 $\pm$ 0.0056} \\
 & Clark $\downarrow$ & 1.6457 $\pm$ 0.0234 $^{\circ}$ & \textbf{1.6439 $\pm$ 0.0237} & 1.7267 $\pm$ 0.0226 $^{\bullet}$ & 1.6495 $\pm$ 0.0230 \\
 & Canberra $\downarrow$ & 3.6912 $\pm$ 0.0552 & \textbf{3.6754 $\pm$ 0.0524} & 3.9466 $\pm$ 0.0586 $^{\bullet}$ & 3.6858 $\pm$ 0.0684 \\
 & Cosine $\uparrow$ & 0.7234 $\pm$ 0.0081 $^{\bullet}$ & 0.7418 $\pm$ 0.0065 $^{\bullet}$ & 0.6560 $\pm$ 0.0067 $^{\bullet}$ & \textbf{0.7496 $\pm$ 0.0078} \\
 & Intersection $\uparrow$ & 0.5840 $\pm$ 0.0071 $^{\bullet}$ & 0.5996 $\pm$ 0.0052 $^{\bullet}$ & 0.5615 $\pm$ 0.0061 $^{\bullet}$ & \textbf{0.6238 $\pm$ 0.0089} \\
\midrule
\multicolumn{2}{l}{Avg. Rank} & 3.50 & 2.13 & 3.27 & 1.10 \\
\bottomrule
\end{tabular}
}
\caption{This table presents the results of the ablation experiments. All results are reported as mean $\pm$ standard deviation over three random splits. The arrow beside the metric indicates the direction for better performance. The optimal value of each row is displayed in bold. we use • (or ◦) to indicate that ERDF is significantly better (or worse) than the corresponding method.}\label{tab:abalation}
\end{table}

\subsection{Ablation study}

To deeply investigate the internal working mechanism of our proposed method, we conduct extensive ablation studies to evaluate the distinct contributions of the two core components of ERDF. Specifically, we compare the full ERDF model with (i) ERDF without the feature enhancement module (w/o fe), (ii) ERDF without the feature reuse module (w/o fr), and (iii) the original DF without either module (w/o fe \& fr). The results are summarized in Table~\ref{tab:abalation}.
Across all datasets and most evaluation metrics, the full ERDF consistently achieves the best performance, demonstrating the complementary benefits of the two modules. Relative to w/o fe \& fr, w/o fe yield noticeable improvements, with average ranks of 2.12. However, its performance ceiling is limited due to the failure to exploit inter-label correlations explicitly. In stark contrast, w/o fr exhibits significant instability. Although it outperforms the baseline from the overall ranking perspective (average rank 3.27), on certain datasets such as SBU, its performance even drops below that of the baseline w/o fe \& fr. This degradation is attributed to the uncontrolled diffusion of noise within the newly generated features, which leads to severe overfitting. This critical phenomenon will be further analyzed in the subsequent subsection. Nevertheless, the superior performance of the full ERDF model vindicates the critical value of the feature enhancement mechanism. It demonstrates that to fully unlock the potential of label correlations without suffering from noise, feature enhancement must be inextricably coupled with feature reuse. Through ablation study it can be seen that the dual-mechanism strategy proposed in this paper offers a robust paradigm for leveraging label correlations within deep forest framwork.

\subsection{Layer-wise Dynamics and Mechanism Analysis}
While the foregoing ablation study quantitatively confirms the complementary nature of the dual-mechanism design, in this section, we leverage a series of visualization experiments on representative datasets to deeply investigate the intrinsic relationship between the two mechanisms proposed in ERDF and the cascade structure of DF. This analysis explicitly elucidates the distinct roles and dynamic evolution of these mechanisms throughout the layer-wise structure.

\paragraph{Interplay between feature enhancement and layer structure}
\begin{figure}[t!] 
    \centering
    
    \begin{subfigure}[b]{0.45\textwidth}
        \centering
        \includegraphics[width=\textwidth]{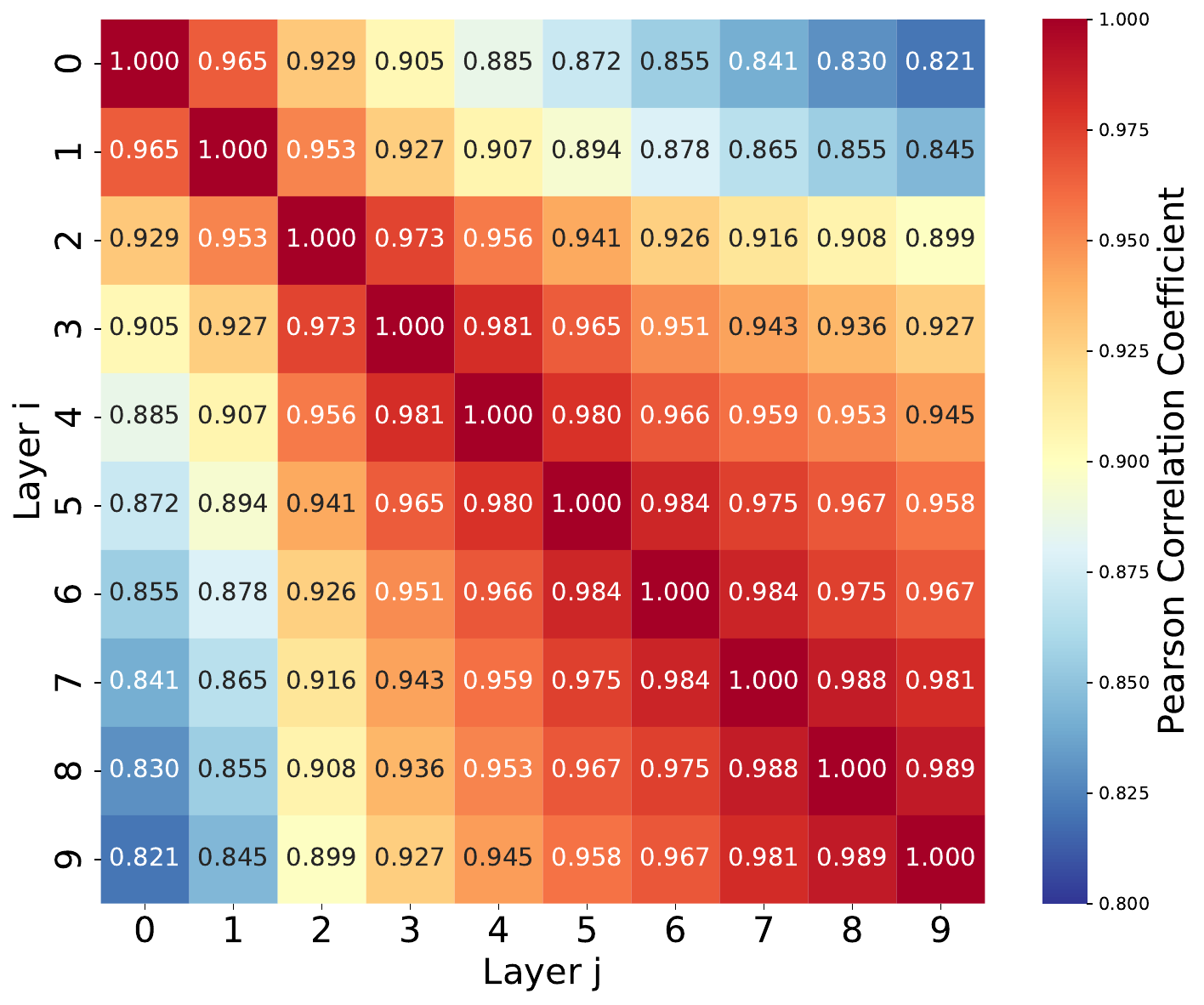} 
        \caption{Movie-heatmap}
        \label{fig:movie_heatmap}
    \end{subfigure}
    \hfill 
    \begin{subfigure}[b]{0.45\textwidth}
        \centering
        \includegraphics[width=\textwidth]{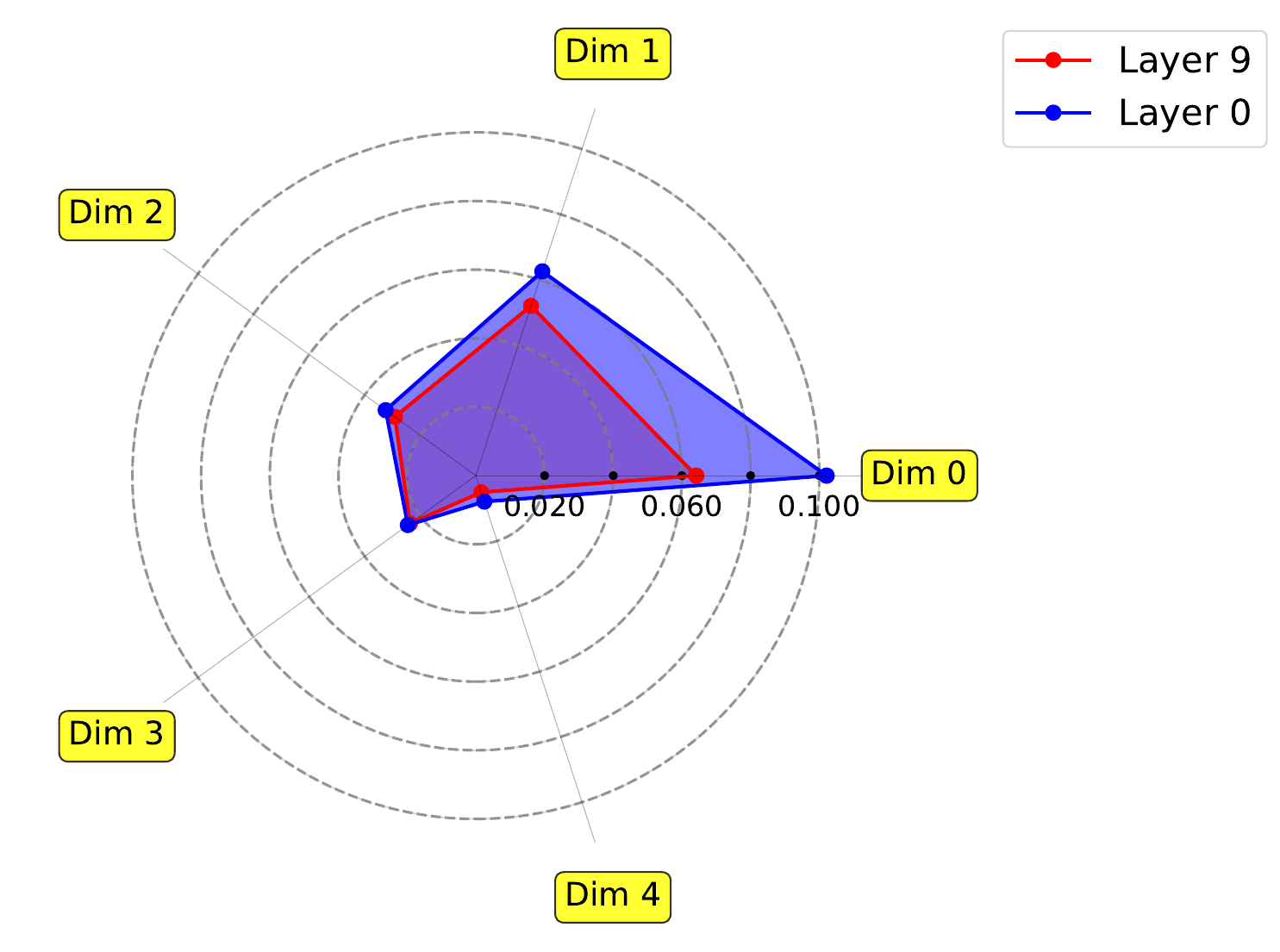}
        \caption{Movie-radar}
        \label{fig:movie_radar}
    \end{subfigure}
    
    \vspace{5pt} 
    
    \begin{subfigure}[b]{0.45\textwidth}
        \centering
        \includegraphics[width=\textwidth]{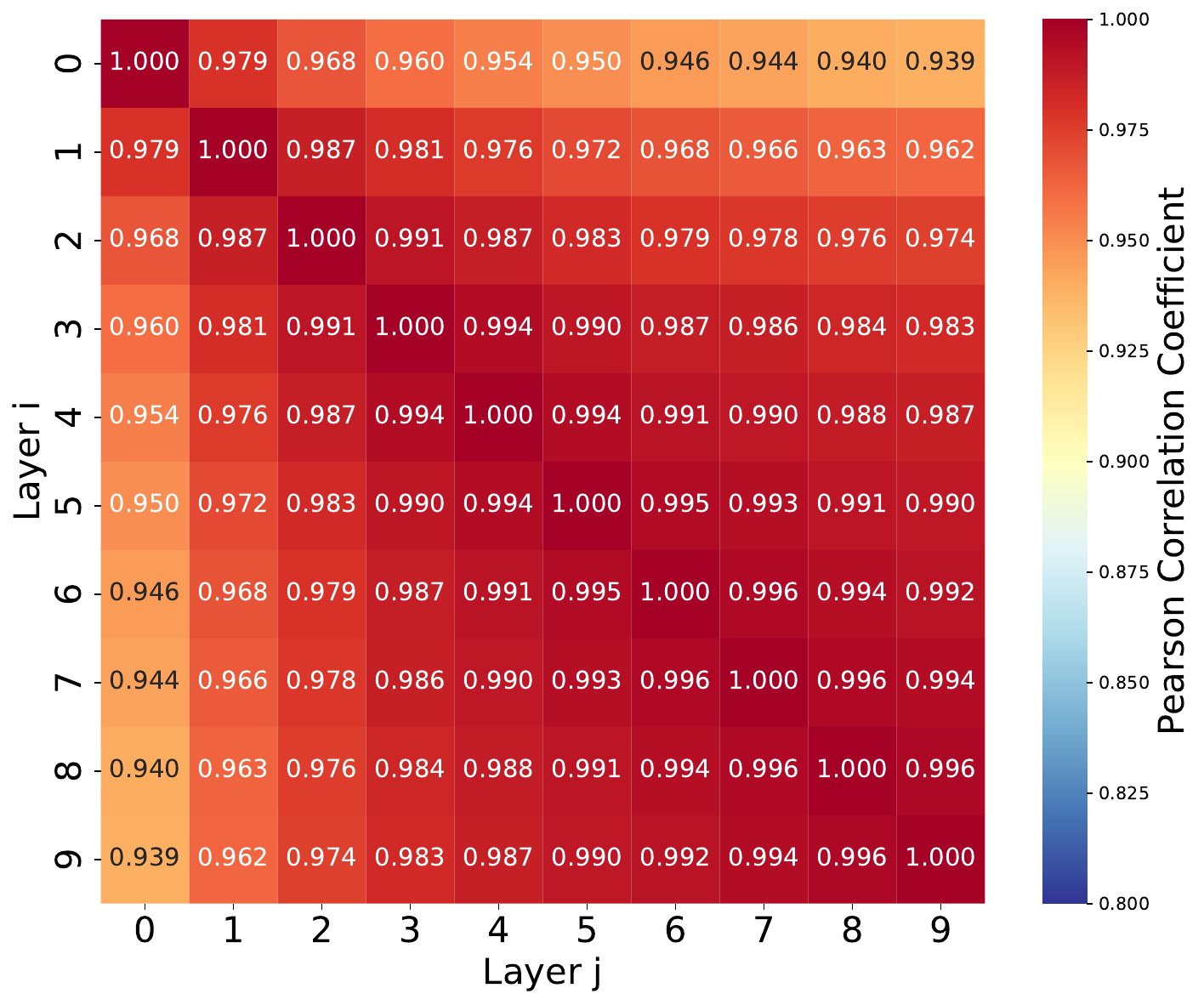}
        \caption{SBU-heatmap}
        \label{fig:SBU_heatmap}
    \end{subfigure}
    \hfill
    \begin{subfigure}[b]{0.45\textwidth}
        \centering
        \includegraphics[width=\textwidth]{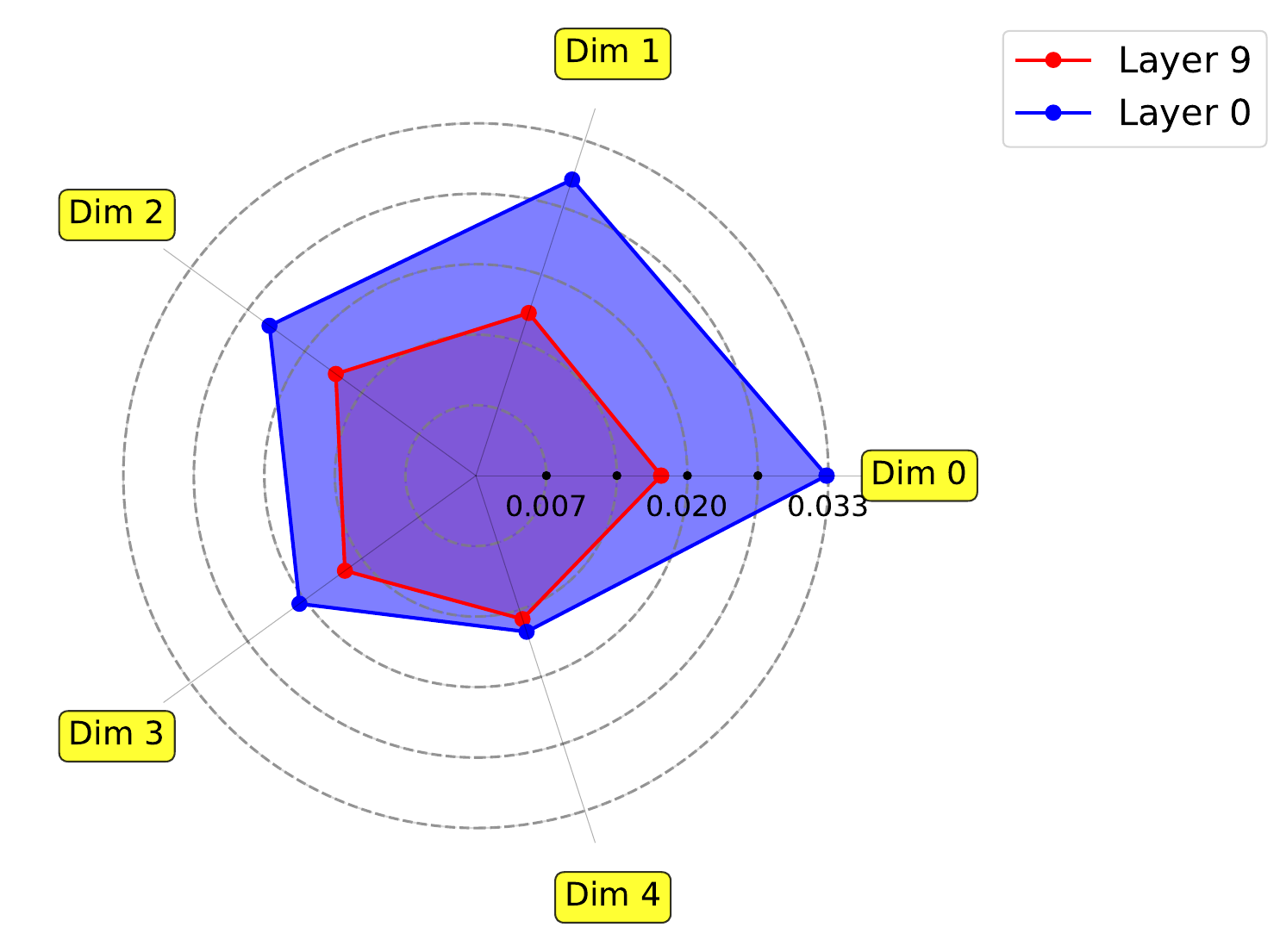}
        \caption{SBU-radar}
        \label{fig:SBU_radar}
    \end{subfigure}
    
    \caption{Visualization of the feature enhancement dynamics across layers on Movie and SBU. Fig.~\ref{fig:movie_heatmap} and \ref{fig:SBU_heatmap} are heatmaps showing the enhanced feature correlation between layers. Fig.~\ref{fig:movie_radar} and \ref{fig:SBU_radar} are radar charts, showing the average error between the predicted values of the enhancers and the ideal values in each dimension. }
    \label{fig:enhancement_analysis}
\end{figure}

To investigate the dynamic evolution of the feature enhancement mechanism throughout the cascading process, we conduct a joint analysis combining inter-layer enhanced feature correlation heatmaps (Fig.~\ref{fig:movie_heatmap} and \ref{fig:SBU_heatmap}) and enhanced feature error radar charts (Fig.~\ref{fig:movie_radar} and ~\ref{fig:SBU_radar}). It is crucial to clarify first that, since the label distribution of the training samples is fixed, the $k$ ideal values of $k$ latent relationship patterns (i.e., the regression targets for the enhancers) extracted via PCA on the label matrix and then calculated by dot product remain invariant across all layers. It is also the reason why the correlation coefficients of the enhanced features between layers are universally high, consistently exceeding 0.8. As the layer depth increases, it is not the regression targets that evolve, but rather the feature representations input into the enhancers. Each layer of DF extracts new features with enhanced discriminability. These are concatenated with the original features to serve as input for the enhancer. This facilitates the learning of an optimal mapping from the feature space to the latent label space, allowing the generated enhanced features to progressively approximate the ideal scores layer by layer. 

Experimental results reveal that feature enhancement acts as two distinct roles during the evolutionary process, contingent upon the dataset characteristics. \textbf{1) Structural Reconstruction and Directional Adjustment.} Taking the Movie dataset as an instance, Fig.~\ref{fig:movie_heatmap} shows that the correlation coefficient between Layer 0 and Layer 9 is only 0.821. This relatively lower inter-layer correlation implies that the model has undergone significant adjustments to the internal structure of the enhanced features as depth increased. Combining this with the radar chart (Fig.~\ref{fig:movie_radar}), it is evident that while the error decreases across all dimensions, this improvement is not significant. This suggests that in complex semantic scenarios, the enhancement mechanism primarily acts as an ``Explorer'', seeking superior representations by continuously reconstructing the direction of the feature space.
\textbf{2) Magnitude Calibration and Numerical Refinement.} Conversely, on the SBU dataset, Fig.~\ref{fig:SBU_heatmap} displays a correlation coefficient of 0.939 between Layer 0 and Layer 9, demonstrating that the distributional trends of the enhanced features maintain high stability across layers. However, the radar chart in Fig.~\ref{fig:SBU_radar} reveals a remarkable error contraction (the red area is significantly smaller than the blue area), with the reduction in the most significant dimension (Dim 0) reaching up to fifty percent. This phenomenon indicates that the initial layers have already captured the correct semantic trends, and the enhancement mechanism in subsequent layers primarily functions as a ``Calibrator''. It focuses on fine-grained magnitude correction of feature values, substantially reducing prediction error while preserving structural stability.

Consequently, the feature enhancement mechanism is not a static procedure but a dynamic optimization process that adaptively performs layer-wise structural reconstruction or numerical calibration depending on the specific statistical characteristics of the dataset.

\paragraph{Interplay between feature reuse and layer structure}
\begin{figure}[t]
    \centering
    \begin{subfigure}[b]{0.49\textwidth} 
        \centering
        \includegraphics[width=\textwidth]{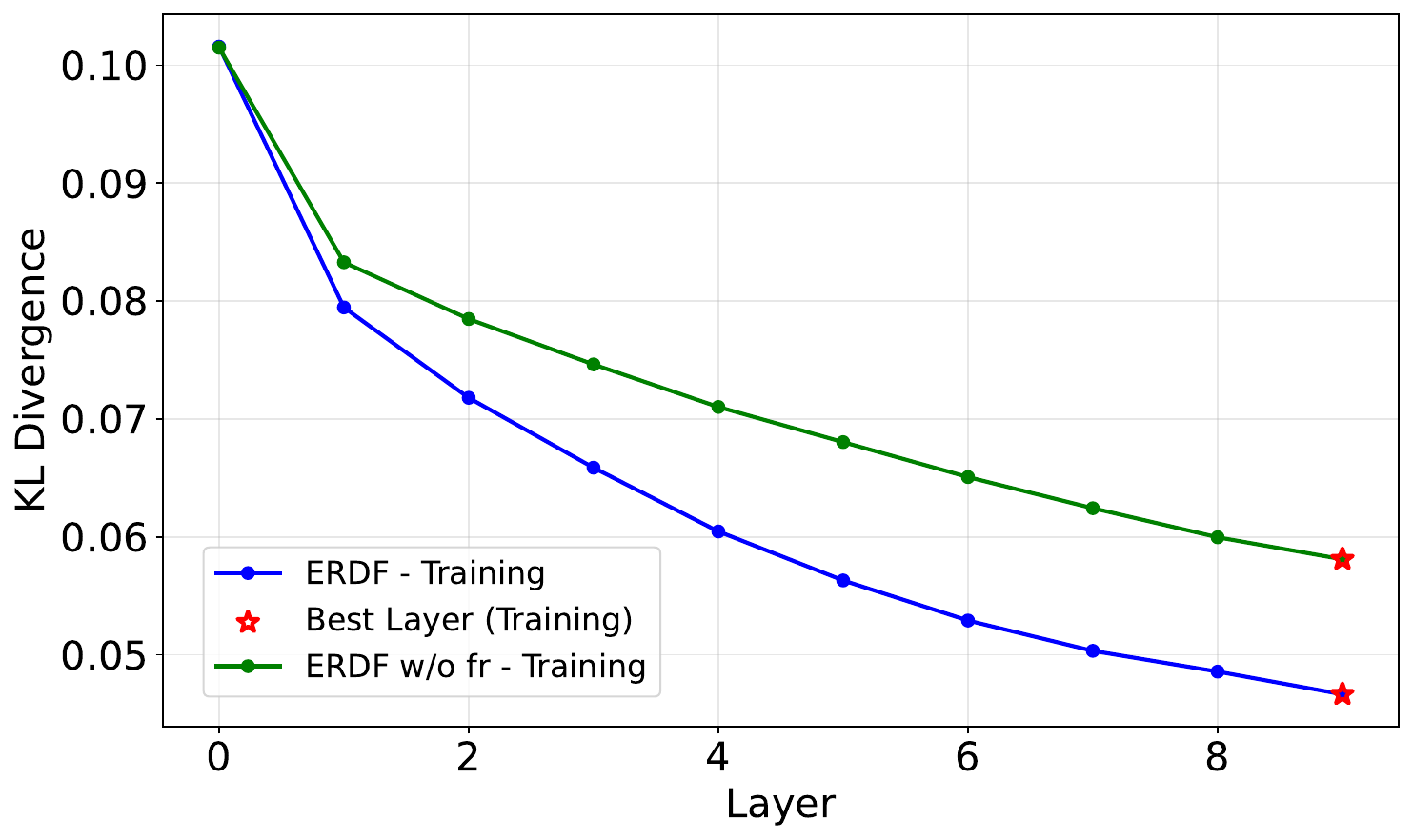} 
        \caption{Movie-training phase} 
        \label{fig:Movie-training}
    \end{subfigure}
    \hfill 
    \begin{subfigure}[b]{0.49\textwidth}
        \centering
        \includegraphics[width=\textwidth]{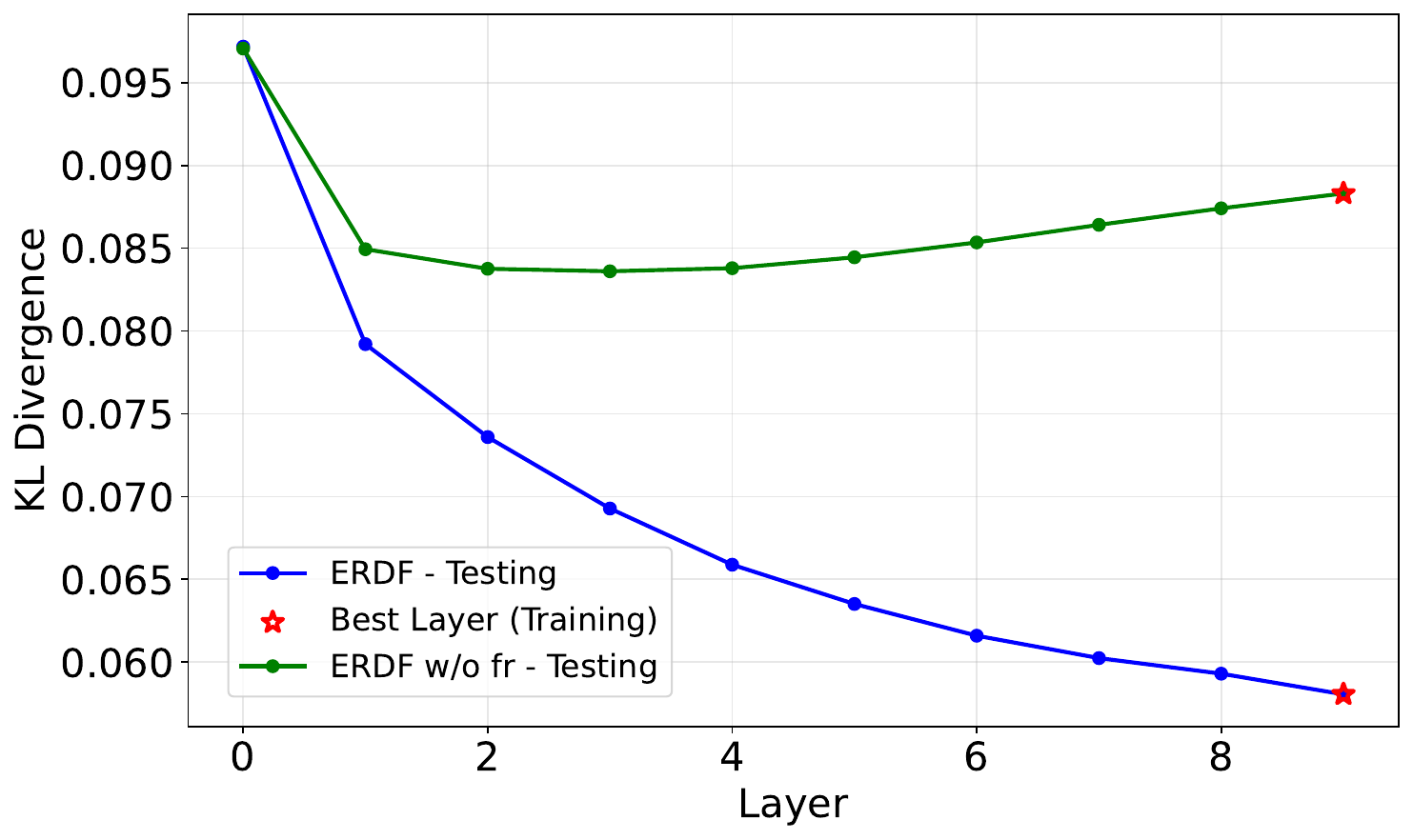}
        \caption{Movie-testing phase}
        \label{fig:Movie-testing}
    \end{subfigure}
    \hfill 
    \begin{subfigure}[b]{0.49\textwidth}
        \centering
        \includegraphics[width=\textwidth]{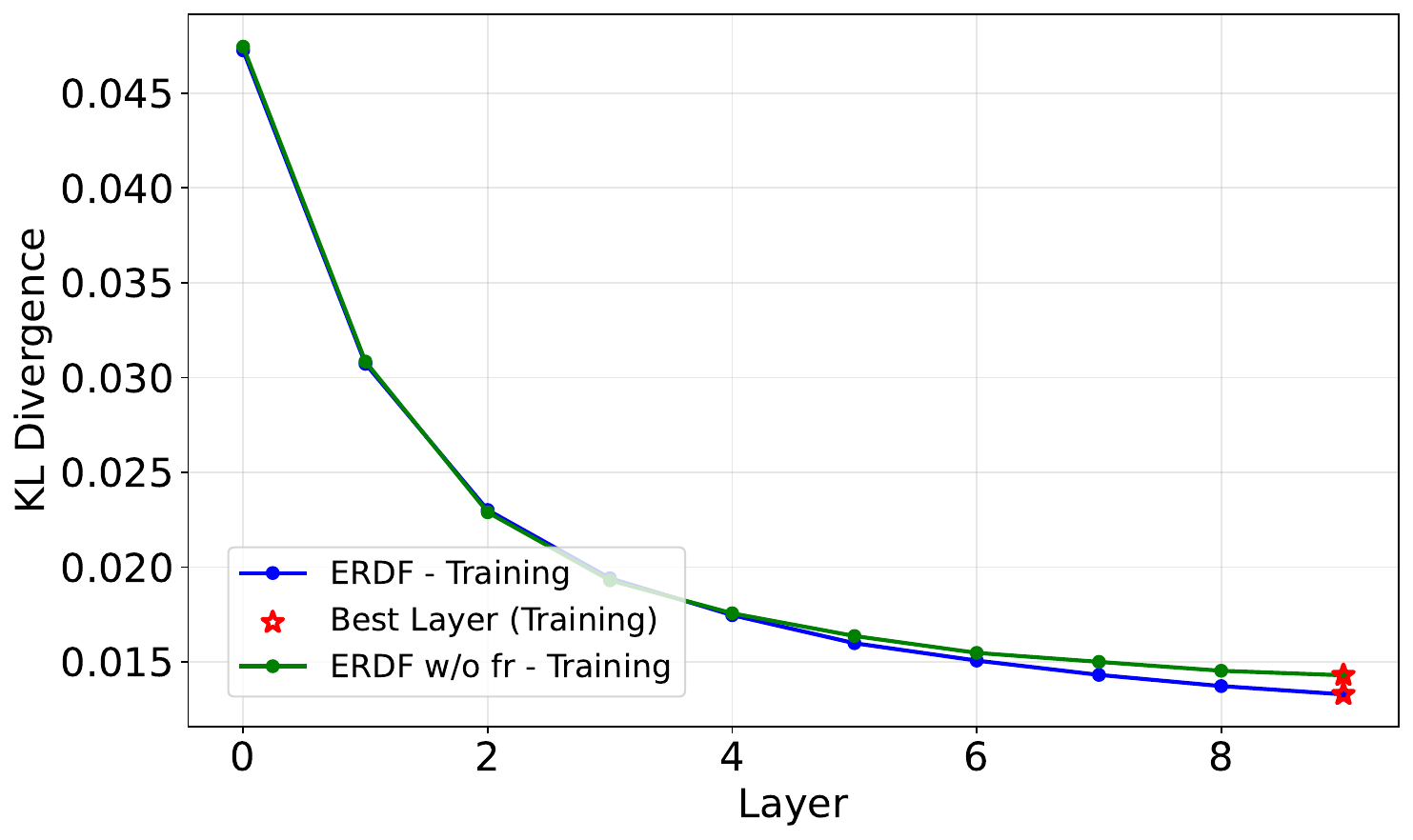}
        \caption{SBU-training phase}
        \label{fig:SBU-training}
    \end{subfigure}
    \hfill 
    \begin{subfigure}[b]{0.49\textwidth}
        \centering
        \includegraphics[width=\textwidth]{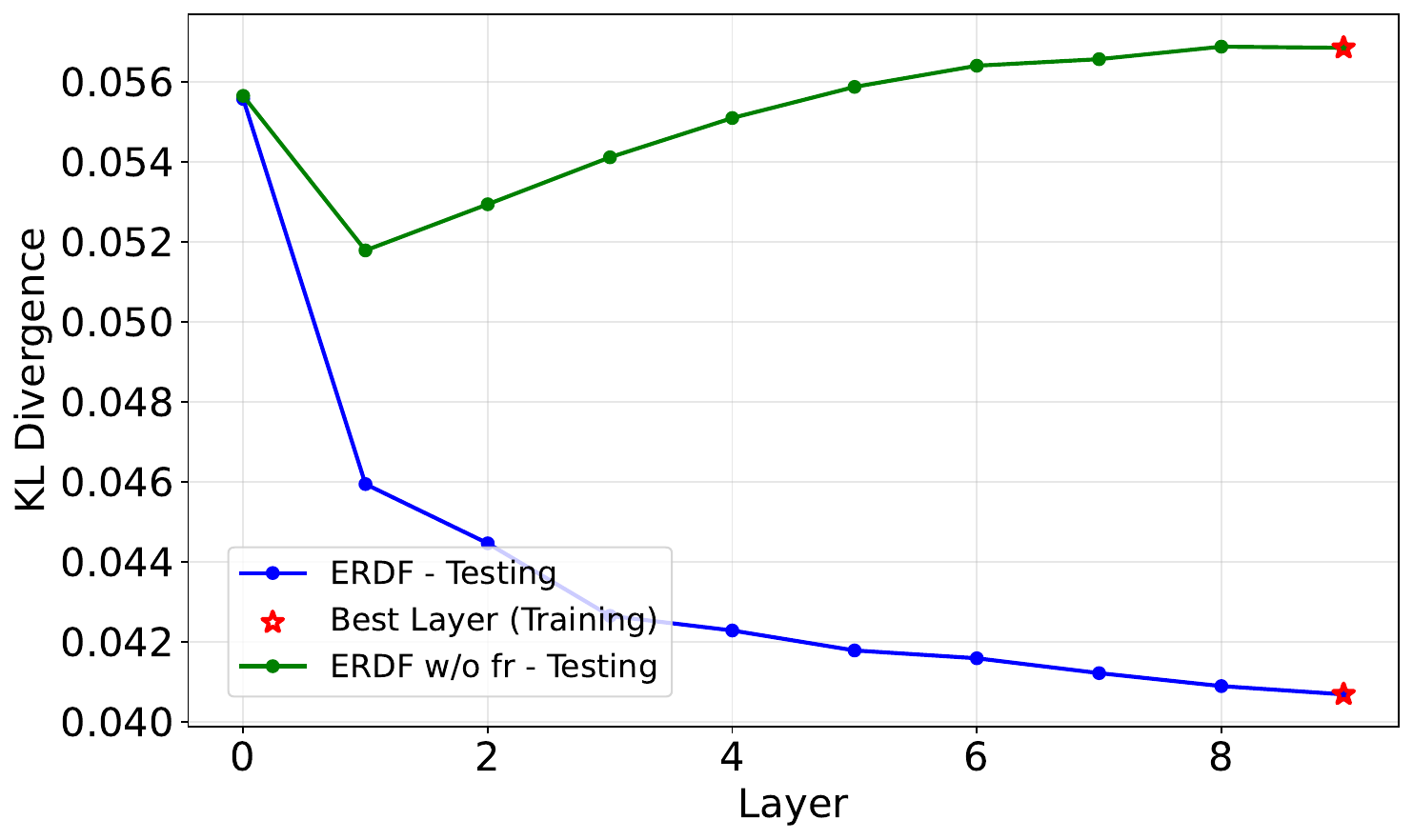}
        \caption{SBU-testing phase}
        \label{fig:SBU-testing}
    \end{subfigure}
    
    \caption{Comparison of KL divergence trajectory between ERDF (blue) and ERDF w/o fr (green) on dataset Movie and SBU. The left column is the training phase, and the right is the testing phase. The red star indicates the optimal layer during the training phase, which is the layer that gives the final prediction in the testing phase.}
    \label{fig:layerkl}
\end{figure}
To validate the robustness of the feature reuse mechanism throughout the cascading process, we compared the layer-wise performance evolution of the complete ERDF model against its variant without the feature reuse mechanism (ERDF w/o fr) on Movie and SBU. As illustrated in Fig.~\ref{fig:layerkl}, we recorded the trajectory of KL divergence during both the training and testing phases.
In the training phase (Fig.~\ref{fig:Movie-training} and \ref{fig:SBU-training}), both curves exhibit a consistent layer-wise downward trend. This indicates that irrespective of the feature reuse mechanism, the cascade architecture of DF possesses potent feature fitting capabilities, enabling it to continuously approximate the training data distribution as the number of layers increases. However, results from the testing phase (Fig.~\ref{fig:Movie-testing} and \ref{fig:SBU-testing}) reveal a fundamental disparity. ERDF w/o fr achieves optimality at the shallow layer, and as the depth further increases, the KL divergence begins to rise rather than decrease. This characteristic ``U-shape'' curve indicates that as the cascade depth grows, the model begins to overfit the noise inherent in the training data. Specifically, due to the feature enhancement mechanism, ERDF introduces higher-dimensional features at each layer compared to traditional DF, which inevitably introduces more potential noise. The noise is amplified layer by layer, ultimately compromising the model's generalization ability.
In contrast, ERDF maintains a continuous downward and stable trend on the testing set, avoiding performance deterioration. This provides compelling evidence that the Feature Reuse mechanism effectively curbs the diffusion of noise within the features. Consequently, it mitigates the overfitting problem in deep cascade structures, ensuring that the model can safely leverage the representational power of deeper structure.

\section{Conclusion}
\label{ch:conclusion}


In this paper, we investigated the problem of label distribution learning and proposed ERDF, a dual-mechanism extension of DF that integrates feature enhancement and measure-aware feature reuse. The proposed framework jointly leverages label correlations to enrich representations while maintaining training stability through adaptive feature reuse. Extensive experiments across multiple benchmark datasets demonstrate that ERDF achieves significant improvements over existing methods, validating both the effectiveness of the two mechanisms.

For future work, two directions appear particularly promising. First, the feature enhancement mechanism is conceptually general and may benefit other tasks that rely on exploiting label dependencies, such as multi-label learning. Exploring these extensions could further broaden the applicability of our approach. Second, the current feature reuse strategy relies on a unified criterion derived from the performance of the first component of the enhanced features. Developing more fine-grained reuse schemes that differentiate among various components of the generated features may further improve stability and performance.

\section*{Declaration of competing interest}
The authors declare that they have no competing interests or financial conflicts to disclose.

\section*{Acknowledgement}
This work was supported by the National Natural Science Foundation of China (No. 62306104, 62441225 and 62572171), Basic Research Program of Jiangsu (No. BK20253011), Hong Kong Scholars Program (No. XJ2024010), Research Grants Council of the Hong Kong Special Administrative Region, China (GRF Project No. CityU11212524), Natural Science Foundation of Jiangsu Province (No. BK20230949), Jiangsu Association for Science and Technology (No. JSTJ2024285), China Postdoctoral Science Foundation (No. 2023TQ0104), and the High Performance Computing Platform of Hohai University.

\bibliography{ref}

\end{document}